# Lifespan Pancreas Morphology for Control vs Type 2 Diabetes using AI on Largescale Clinical Imaging

Lucas W. Remedios[a,*], Chloe Cho[e], Trent M. Schwartz[b], Dingjie Su[a], Gaurav Rudravaram[b], Chenyu Gao[b], Aravind R. Krishnan[b], Adam M. Saunders[b], Michael E. Kim[a], Shunxing Bao[b], Thomas A. Lasko[a,m], Alvin C. Powers[j,k,l], Bennett A. Landman[a,b,e], John Virostko[f,g,h,i]

[a]Vanderbilt University, Department of Computer Science, Nashville, USA; [b]Vanderbilt University, Department of Electrical and Computer Engineering, Nashville, USA; [e]Vanderbilt University, Department of Biomedical Engineering, Nashville, USA; [f]University of Texas at Austin, Department of Diagnostic Medicine, Dell Medical School, Austin, USA; [g]University of Texas at Austin, Livestrong Cancer Institutes, Dell Medical School, Austin, USA; [h]University of Texas at Austin, Austin, Department of Oncology, Dell Medical School, USA; [i]University of Texas at Austin, Oden Institute for Computational Engineering and Sciences, Austin, USA; [j]Vanderbilt University Medical Center, Department of Medicine, Division of Diabetes, Endocrinology, and Metabolism, Nashville, USA; [k]Vanderbilt University, Department of Molecular Physiology and Biophysics, Nashville, USA; [l]VA Tennessee Valley Healthcare System, Nashville, USA; [m]Vanderbilt University, Department of Biomedical Informatics, Nashville, USA

**Abstract**

**Purpose:** Understanding how pancreas size and shape change with normal aging is critical for establishing a baseline to detect deviations in type 2 diabetes and other pancreatic disease. We measure pancreas size and shape using morphological measurements from early development through aging (ages 0-90). Our goals are to 1) identify reliable clinical imaging modalities for artificial intelligence (AI) based pancreas measurement, 2) establish normative morphological aging trends, and 3) detect potential deviations in type 2 diabetes.

**Approach:** We analyzed a clinically acquired dataset of 2533 patients imaged with abdominal computed tomography (CT) or magnetic resonance imaging (MRI). The patients did not have cancer, pancreas pathology, sepsis, or trauma. We resampled the scans to 3mm isotropic resolution, segmented the pancreas using automated methods, and extracted 13 morphological pancreas features across the lifespan. First, we assessed pancreas volume trajectories in 1858 control patients across contrast CT, non-contrast CT, and MRI to determine which modalities provide consistent lifespan trends. Second, we characterized distributions of normative morphological patterns stratified by age group and sex. Third, we used covariate-adjusted generative additive models for location, scale, and shape (GAMLSS) regression to model pancreas morphology trends in 1350 patients matched for age, sex, and type 2 diabetes status to identify any deviations from normative aging associated with type 2 diabetes.

**Results:** When adjusting for confounders, the aging trends for 10 of 13 morphological features were significantly different between patients with type 2 diabetes and non-diabetic controls ($p < 0.05$ after multiple comparisons corrections). Additionally, MRI appeared to yield different pancreas measurements than CT using our AI-based method.

**Conclusions:** We provide lifespan trends demonstrating that the size and shape of the pancreas is altered in type 2 diabetes using 675 control patients and 675 diabetes patients. Moreover, our findings reinforce that the pancreas is smaller in type 2 diabetes. Additionally, we contribute a reference of lifespan pancreas morphology from a large cohort of non-diabetic control patients in a clinical setting.

*Lucas W. Remedios, E-mail: lucas.w.remedios@vanderbilt.edu

# 1 Introduction

Substantial efforts have been made to measure the morphology of the structural aging of organs, such as the brain, which helps distinguish between atrophy caused by normal aging and atrophy caused by disease[1]. Defining an organ's normative morphological aging allows investigation into how disease alters this trajectory, offering insight into the disease process. As with the brain, the pancreas also undergoes changes with age[2] (Figure 1).

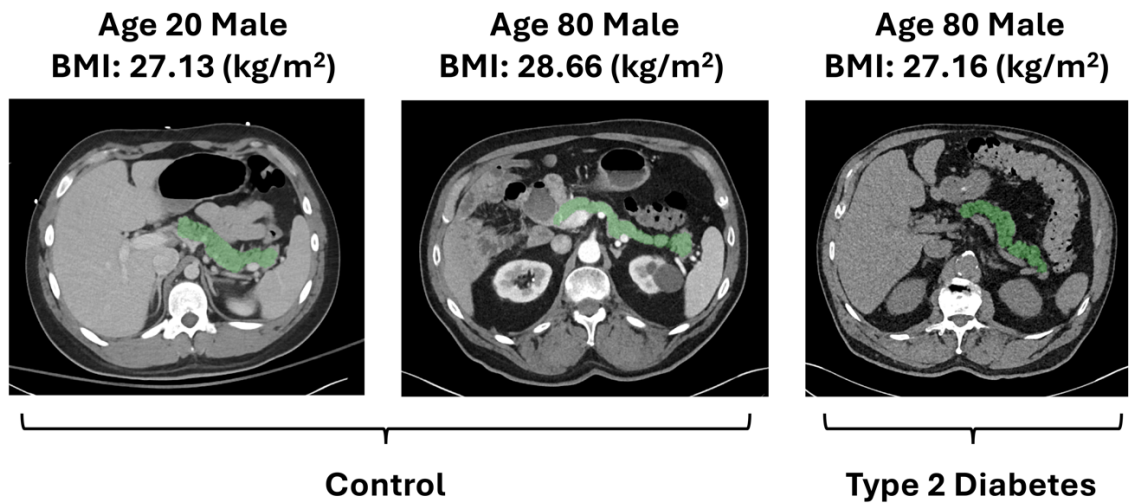

**Fig. 1** The pancreas undergoes structural changes with age, including atrophy and fat infiltration. While population-level pancreas volume and fat content have been examined across the aging process[3], there remains a knowledge gap in understanding age-related changes in the pancreas across a broader set of morphological measurements. Moreover, type 2 diabetes may cause changes in pancreas morphology that differ from normal aging. The two scans on the left illustrate age-related appearance differences in non-diabetic patients but are not from the same patient. The rightmost scan shows the pancreas from an elderly patient with type 2 diabetes. Understanding pancreas variation in normal aging is critical for understanding differences in type 2 diabetes. Any potential differences in the aging trends of the pancreas in type 2 diabetes may not necessarily be linear or smooth.

As the pancreas ages, it changes in size, shape, fat content, and duct structure, along with increasing fibrosis[4,5,6,7,8,9]. Separately from aging, pancreas morphology is influenced by sex and body composition[11,12]. Changes in pancreas morphology have also been linked to diseases including diabetes[10].

Pancreas volume is a critical measure of pancreas morphology. Nearly 20 years ago, Saisho et al. measured pancreas volume across the lifespan, where the pancreas was segmented (outlined by hand) from computed tomography (CT)[3]. Today, artificial intelligence (AI) enables faster, fully-automated segmentation of the pancreas[11]. Pancreas volume declines with age and is partially replaced by ectopic fat[3,13,14]. In type 2 diabetes, the pancreas has been found to be smaller than control subjects, and has a serrated edge[15,16.] In Saisho et al., the pancreas was determined to be smaller in subjects with type 2 diabetes in a large dataset matched for age, sex, and body mass index (BMI)[3] — this finding aligned with previous studies[17–21], but also differed from several other studies that found no volume difference in type 2 diabetes from control subjects[22–25].

Beyond volume, deep characterization of pancreas morphology may lead to insights on disease states. The diameter of the pancreas has been identified as an important feature of pancreatic health[15]. Additionally, pancreas shape features have been extracted on magnetic resonance imaging (MRI) to study nearly 4000 subjects[26].

In this work, we measure the pancreas from a large clinical dataset via AI-driven pancreas segmentation (Figure 2). We comment on the consistency of clinical CT and MRI for volume measurements, and present lifespan trends on normative aging with 13 morphological features of the pancreas. Further, we test for differences in structural aging between control and type 2 diabetes in a dataset matched on sex and age spanning ages 20 to 90. Rather than correcting for body size through division, we model aging trends while accounting for the covariates (age, sex, weight, and diabetes status).

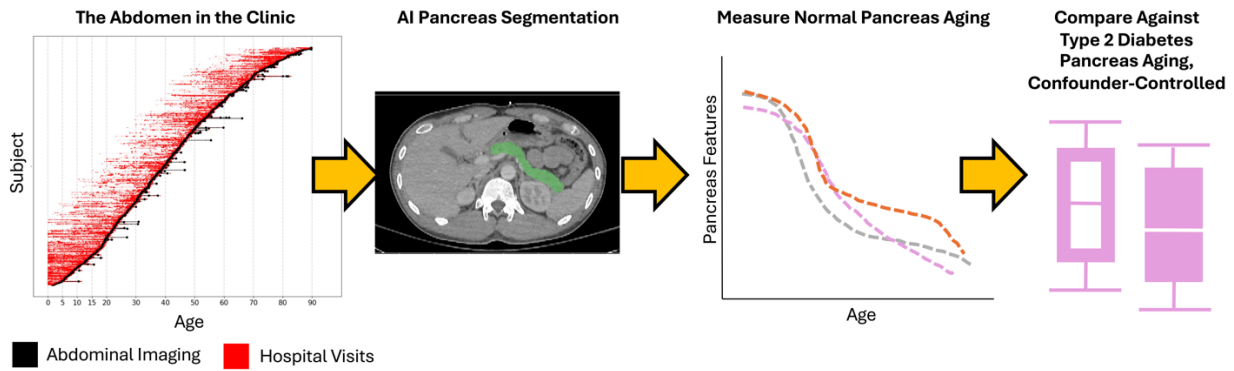

**Fig 2** We leverage over 2500 clinical abdominal scans (CT or MRI) from control and type 2 diabetes patients. Using the AI tool TotalSegmentator[48,49], we automatically segment the pancreas. We then use PyRadiomics[50] to extract 13 morphological measurements across the lifespan to assess whether pancreas aging differs in type 2 diabetes. We control for age, sex, and weight effects through both matching and modeling.

## 2  Methods

*2.1 Data*

Our dataset was retrieved in deidentified form from Vanderbilt University Medical Center under IRB #241494. After full processing, detailed below, there were 2533 patients, each with one abdominal scan (CT or MRI), and each with a single diabetes label, either control or type 2 diabetes. For compatibility with downstream processing, the images were converted from DICOM to NIfTI format via dcm2niix[51].

*2.2 AI segmentation & Feature Extraction*

The scans were converted into LAS orientation, cropped between the L5 and T7 vertebrae (TotalSegmentator[52] vertebrae segmentation), and resampled to 3mm isotropic resolution. The pancreas and other abdominal organs were segmented with TotalSegmentator[52,53] version 2.8, using the CT model for CT scans, and the MRI model for MRI scans. Thirteen morphological features were extracted from the binary NIfTI pancreas segmentations using PyRadiomics[50] version 3.1. The features were: volume, surface area, surface area to volume ratio, elongation, flatness, sphericity, major axis length, minor axis length, least axis length, maximum 3D diameter, maximum 2D diameter column, maximum 2D diameter row, and maximum 2D diameter slice.

*2.3 Quality Control of Medical Images & Segmentations*

To ensure data quality, all images and organ segmentations were manually inspected with a high-throughput visualization tool[54]. In some cases, the conversion from DICOM to NIfTI format failed and corrupted image metadata. In these failure cases, the images were warped to varying degrees in any combination of the axial, sagittal, and coronal axes. Additionally, undesirable cases contained organ segmentations with more than one contiguous volume, or that touched the edge of the image volume, indicating the organ extended beyond the field-of-view. These scans were excluded, with the warping identified via out-of-distribution examples of the ratio of the faces on a bounding box of the liver segmentation.

A subsequent manual inspection of the images and segmentations revealed that the warped scans persisted in the dataset. After examining that the variability of liver volume within an imaging session could identify sessions with warped scans, we fit a polynomial across the lifespan to each of these abdominal organs: pancreas, liver, spleen, left kidney, and right kidney. Within each session, we selected the scan with the volumes closest to the lifespan polynomials, ranked with even weighting across the five organs. This approach made a tradeoff: we removed erroneous data, but at the same time potentially limited the true representation of biological variability in the dataset. Another round of manual inspection of the images led to the exclusion of all the warped images in the dataset.

*2.4 Diabetes Label Assignment*

To assign diabetes labels to the patients, we used their health records up to 1 year after the selected scan. International classification of disease (ICD) codes and derived PhecodeX[55] were used to determine control patients as those without any diabetes events, and patients with type 2 diabetes as those with at least one type 2 diabetes event and no type 1 diabetes events. Where available, A1C measurements were used to validate that the control patients did not have diabetes. Using diagnosis information from ICD and PhecodeX, as well as procedure information from their current procedural terminology (CPT) and procedure wide association (ProWAS) codes, we excluded patients with cancer, pancreas pathology, and sepsis[56,57]. Pancreas pathology was identified by filtering ICD and PhecodeX records for the word segment "panc". We also excluded imaging associated with trauma events to avoid scans with acute structural and functional alterations (e.g., hemorrhage, edema).

*2.5 Automatic Contrast Phase Labeling of CT*

CT scans were automatically labeled according to contrast phase using TotalSegmentator's contrast phase prediction tool[52]. These labels were inspected and determined to be accurate for binary contrast labeling (contrast CT or non-contrast CT). The final binary phase classification was manually assured for each CT through visual inspection of the scans using AutoQA[54].

*2.6 Pancreas Measurement Consistency across CT & MRI*

Before analyzing anatomical trends, we needed to select which imaging modalities (CT or MRI) in the clinical dataset consistently measured the pancreas. We chose to use pancreas volume, pancreas volume index (volume divided by patient weight), and BMI of healthy controls to assess this. Based on observation of lifespan polynomials fit with 95% confidence intervals, we selected the subset of modalities to use in subsequent analyses.

*2.7 Lifespan Pancreas Morphology in the Clinic by Sex*

On our selected modalities, we created a lifespan reference of normative pancreas morphology by creating boxplots to visualize the distributions of each uncorrected pancreas morphological feature by sex and age group as in Saisho et al.[3] These data were separated by sex but were not matched.

*2.8 Modeling Pancreas Aging with and without Diabetes*

We modeled each pancreas feature across the adult lifespan (ages 20–90) using a single generalized additive model for location, scale, and shape (GAMLSS[58]), jointly incorporating both male and female, as well as individuals with and without type 2 diabetes. The location ($\mu$) was modeled as:

$$\mu = \beta_0 + f(\text{age}) + \beta_1 \text{ diabetes} + \beta_2 \text{ sex} + \beta_3 \text{ weight} \qquad (1)$$

Here, $\beta_0$ is the intercept, and $\beta_1$, $\beta_2$, and $\beta_3$ are coefficients for type 2 diabetes status, sex, and weight respectively. The term $f(\text{age})$ represents a smooth, nonlinear function of age, implemented using penalized B-splines. We selected this GAMLSS model to flexibly capture the nonlinear nature of biological aging. In contrast, diabetes, sex, and weight were modeled linearly under the assumption that they introduce global shifts. The same covariates were included in the model for the scale ($\sigma$), and we used the Box-Cox Cole and Green (BCCG) distribution to handle skewness in the feature distributions. Although GAMLSS does not include an explicit residual term like in traditional linear models, the progression-related variability (i.e., error) is captured through the distribution parameters. The GAMLSS approach enables learning distributions, rather than just the mean, as a function of the input variables.

Although one model was fit per pancreas feature, we visualized the fitted curves separately for male and female to highlight sex-specific differences. For each sex, weight was fixed to the average weight across all patients of that sex (both type 2 diabetes and control), and we conditioned the visualization on type 2 diabetes status to isolate its impact on pancreas aging. Because we modeled type 2 diabetes diagnosis as a linear relationship, we obtained a single *p*-value for the differences between the type 2 diabetes and control groups. These *p*-values were corrected using the Benjamini–Hochberg False Discovery Rate (FDR) to account for the 13 features tested. After FDR correction, $p < 0.05$ was considered significant.

*2.9 Use of Generative AI in Manuscript*

Generative AI (ChatGPT4o) was employed to assist with drafting and content refinement throughout this research. All core intellectual content and insights originated from the authors' independent scholarly work. All AI-generated material passed through careful author review and revision to ensure alignment with the study's objectives and to maintain academic rigor.

# 3    Results

*3.1 Comparison of Clinical Imaging Modality for Measuring the Pancreas*

In Figure 3, we observed for agreement between contrast CT, non-contrast CT, and MRI for measuring the pancreas via lifespan trends (polynomials). We selected pancreas volume as the main feature, since population-level reference ranges are provided by Saisho et al.[3]. We used an unmatched subset of the dataset consisting of all 1858 control patients. We observed that MRI tended to yield smaller pancreas measurements than CT. To account for differences in weight distributions, we also assessed the pancreas volume index (pancreas volume divided by patient weight), where the pattern was still observed. We additionally checked the distribution of BMI across modalities, which was comparable. Based on this observed incongruency between image modalities, we cautiously excluded automated pancreas segmentation derived from MRI from our subsequent analyses and instead focused on the larger group of CT images.

*3.2 Lifespan Pancreas Morphology Shows Aging Trends*

In Figure 4, we present uncorrected lifespan morphology of the pancreas in an unmatched dataset of 1775 nondiabetic control patients with CT imaging of the pancreas. Our results reproduce the population volume trends from Saisho et al.[3] Across size metrics (axis lengths and diameters), the female pancreas is generally smaller in adulthood.

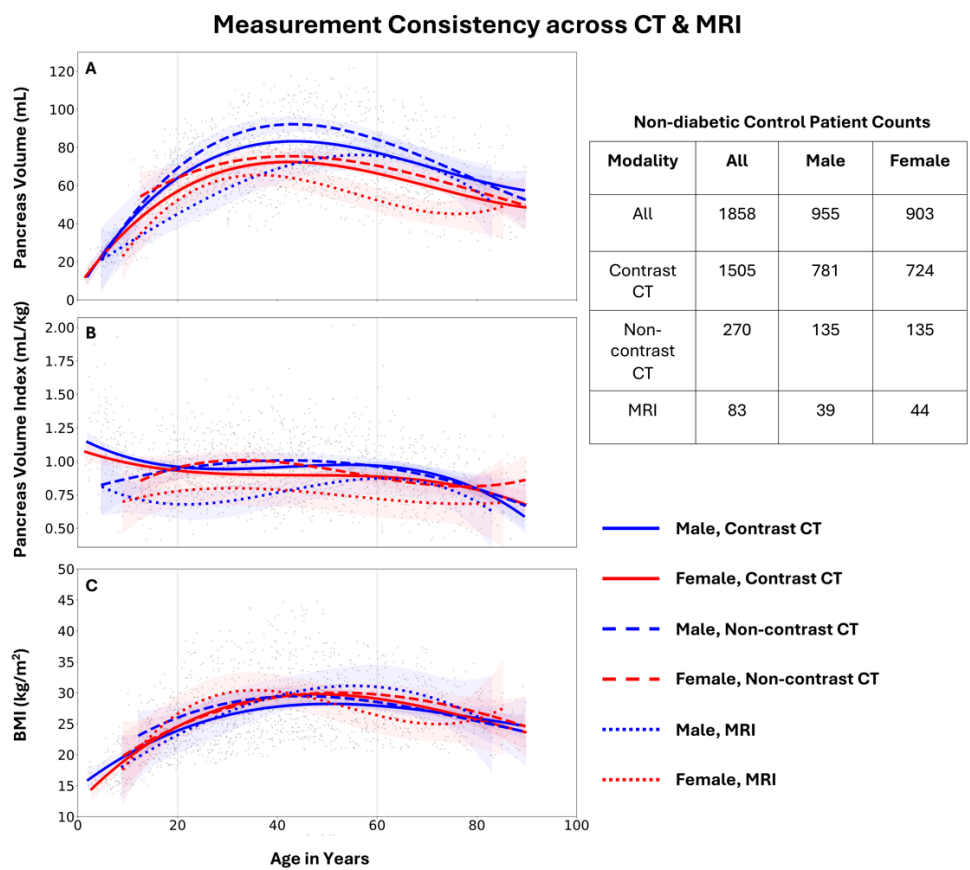

**Fig 3** We investigate measurement consistency between CT and MRI using lifespan polynomials. With our automatic AI approach, MRI appears to measure a smaller pancreas volume than CT for most age groups (A). The observable difference between the MRI and CT trends becomes more evident when correcting for body size by dividing pancreas volume by patient weight (B). BMI is comparable across the modalities, which further implies that observed reduced pancreas measurements in MRI are likely not caused by smaller body size (C). These trends are from non-diabetic control patients. Given the observed discrepancy in automated pancreas volume measures between MRI and CT, we cautiously exclude the smaller cohort of MRI measurements from further analyses and proceed with CT measurements.

*3.3 Matched Dataset for Assessing Control vs Type 2 Diabetes*

In Figure 5, we show the distribution of matched patients by age, sex, and diabetes status. These 1350 patients allow a large sample size for measuring potential adulthood pancreas aging differences in type 2 diabetes, with most of the data being available from age 40 to 70.

*3.4 Shift in Diabetes*

In Figure 6, we show that when accounting for covariates, 10 of 13 morphological measurements of the pancreas were significantly different ($p < 0.05$ after multiple comparisons corrections on the linear type 2 diabetes parameter). On the statistically significant features, there were small shifts between the type 2 diabetes and control curves, indicating that type 2 diabetes is associated with measurable changes in pancreas morphology throughout adulthood. These differences persisted even after both matching and modeling the data to account for age, sex, and body weight using GAMLSS. Significant differences were observed across most of the anatomical measurements, including pancreas volume, surface area, surface-to-volume ratio, elongation, flatness, sphericity, and multiple axis and diameter-based measurements. We also observed sex-specific trends, with shifts between male and female trajectories. The findings in Figure 6 support that the type 2 diabetes pancreas is smaller than control and has altered morphology throughout aging in adulthood, however there is a large amount of overlap in the distributions between type 2 diabetes and control. In Figure 7, we present complementary numerical details for the type 2 diabetes coefficient from the $\mu$ of the GAMLSS model.

*3.5 Visualizing the Aging Pancreas*

In Figure 8, we present a scaled lifespan visualization of the pancreas across sex and type 2 diabetes status. We show examples with sampled near median volume from each group and use maximum intensity projections to visualize the entire pancreas segmentation from an axial view. We overlay the pancreas on maximum intensity projections of the body and other abdominal organs for context. Earlier, our quantitative methods revealed significant differences in the aging pancreas in type 2 diabetes, but with large overlap in distributions between type 2 diabetes and control. In Figure 8, differences in pancreas visualization are not readily discernible in our maximum intensity projection visualization at 3mm isotropic resolution.

## 4   Discussion

In this study, we characterized age-related changes in pancreas size and shape using a large, clinically acquired dataset and AI-based segmentation across CT and MRI modalities. We found that pancreas morphology ages differently in type 2 diabetes across numerous measures of size and shape. Our modality comparison led to observed differences between MRI and CT in AI-based pancreas measurement, emphasizing the need for modality-specific baselines.

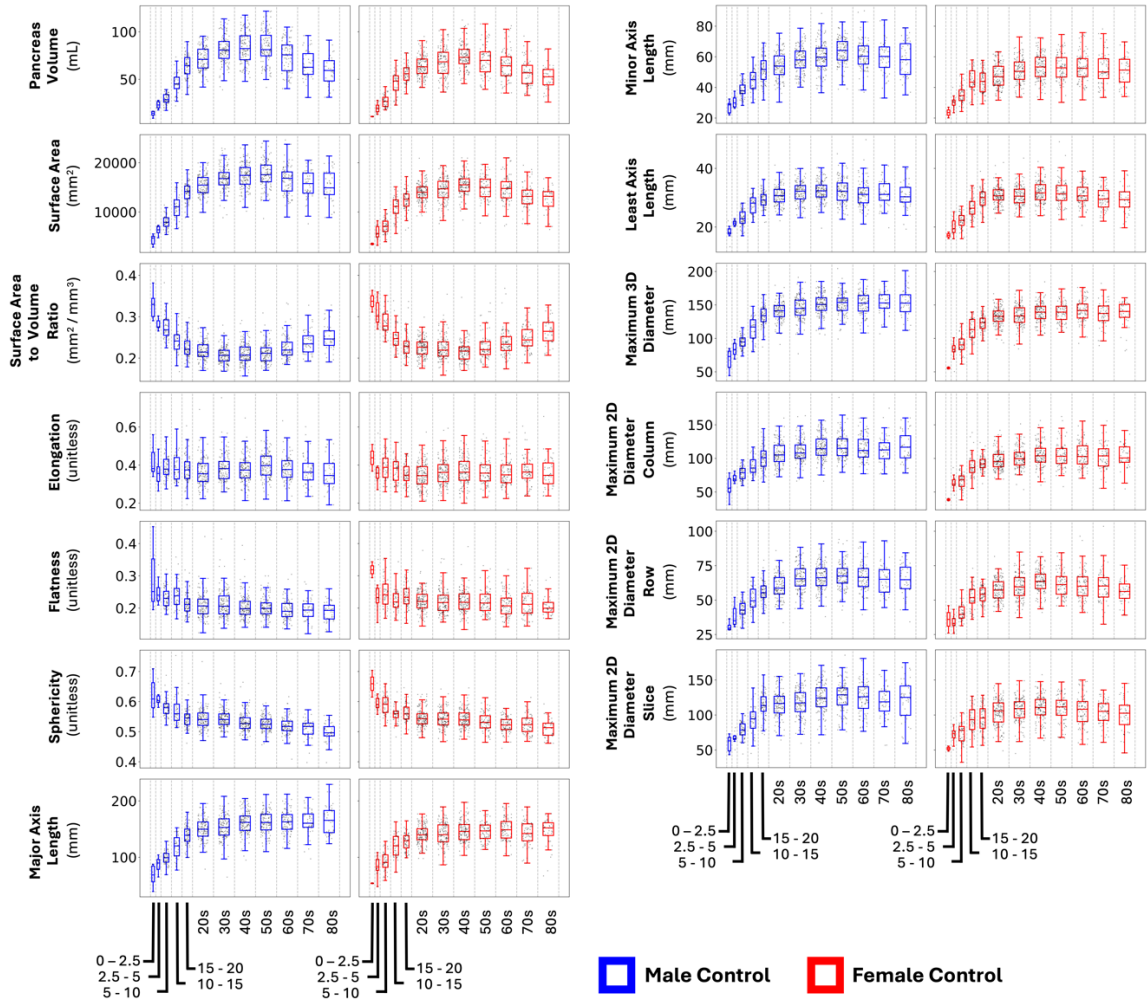

**Fig 4** From a population of 1775 non-diabetic control patients with CT scans, we illustrate how pancreas size and shape change with age and sex across 13 morphological measurements. We reproduce findings from Saisho et al. on pancreas volume[3]. These distributions visualize population spread of pancreas measurements across age groups.

Measuring the pancreas from medical imaging is commonly performed on both CT and MRI in research settings. In clinical populations, CT is more prevalent, however MRI is commonly performed in research studies to avoid radiation exposure[27]. In datasets of clinically acquired abdominal CT scans, images reflect real-world variability in image quality and acquisition protocols, which may influence pancreas appearance and measurement. Several previous studies have inspected the pancreas from CT and/or MRI[28,29,30,31,32,33,34].

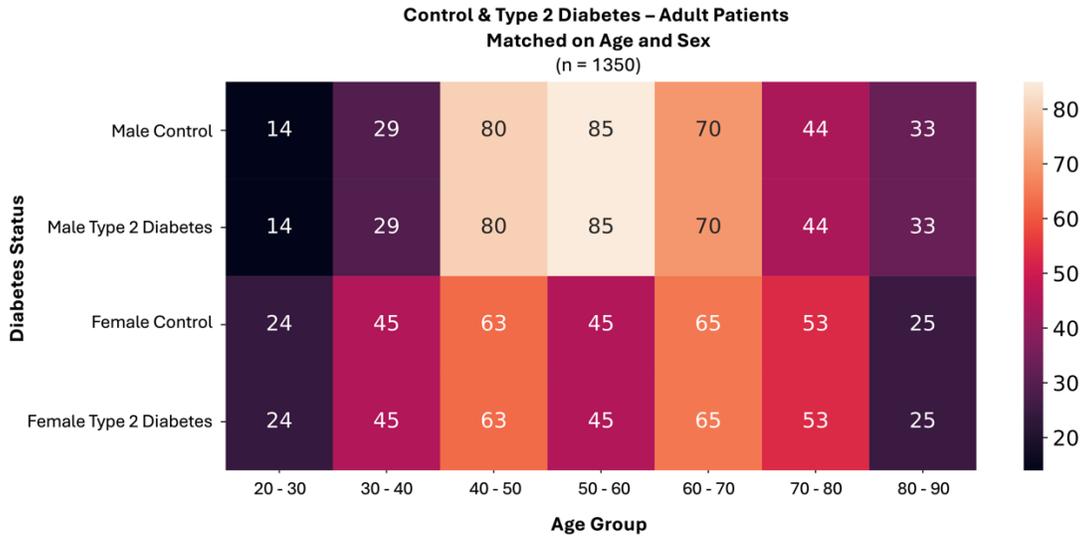

**Fig 5** We used a matched subset of the data with 1350 patients to mitigate the confounding effects of sex and age before assessing whether pancreas morphology differs in type 2 diabetes. Body size was not matched but was addressed later through modeling. Each cell provides the number of patients, colored by the color bar, in each diabetes/sex and age group.

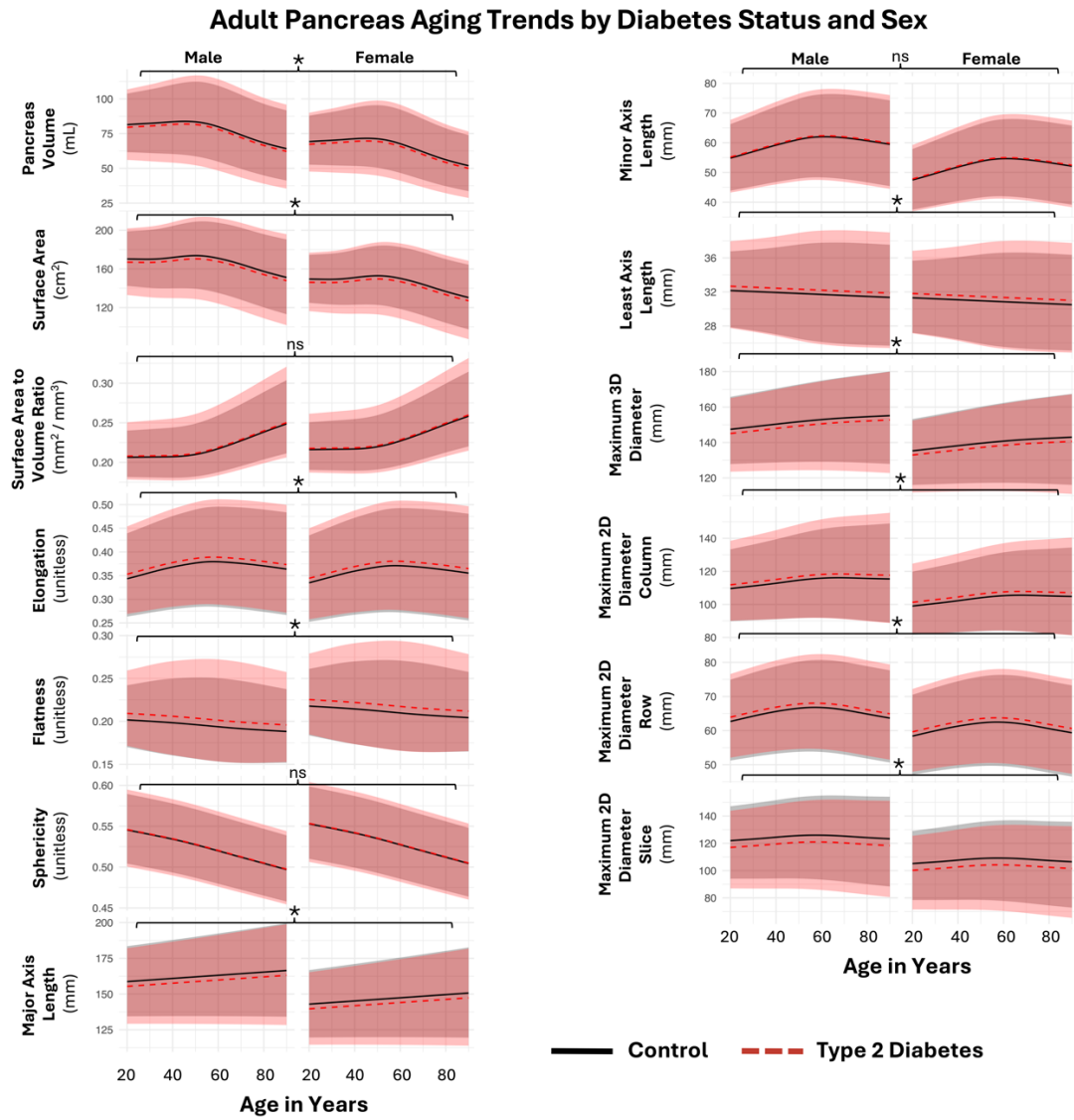

**Fig 6** In type 2 diabetes, the pancreas is smaller and has an altered shape with our GAMLSS regression that models diabetes diagnosis with a linear term. Across these 13 morphological measurements, 10 of the aging trends were significantly different in diabetes, with $p<0.05$ after multiple comparisons correction denoted with *. In these plots, we show the 50th percentile of the learned distributions, with the range from the 5th to 95th centiles shaded, holding other covariates constant. While we detect statistical significance via the linear type 2 diabetes parameter across most of the metrics, the distributions from the patients with type 2 diabetes are similar to the non-diabetic controls.

| Pancreas Features | Coefficient | Standard Error | p-value | FDR-corrected p-value |
|---|---|---|---|---|
| Pancreas Volume | -1.92E+00 | 9.00E-01 | 3.27E-02 | 4.25E-02 |
| Surface Area | -3.32E+00 | 1.27E+00 | 9.31E-03 | 2.02E-02 |
| Surface Area to Volume Ratio | 1.50E-03 | 1.33E-03 | 2.60E-01 | 3.08E-01 |
| Elongation | 9.56E-03 | 3.89E-03 | 1.40E-02 | 2.03E-02 |
| Flatness | 7.59E-03 | 1.81E-03 | 3.04E-05 | 1.97E-04 |
| Sphericity | 5.19E-04 | 1.53E-03 | 7.34E-01 | 7.34E-01 |
| Major Axis Length | -3.34E+00 | 1.02E+00 | 1.07E-03 | 4.65E-03 |
| Minor Axis Length | 3.39E-01 | 4.77E-01 | 4.77E-01 | 5.17E-01 |
| Least Axis Length | 5.05E-01 | 2.03E-01 | 1.31E-02 | 2.03E-02 |
| Maximum 3D Diameter | -2.40E+00 | 8.17E-01 | 3.36E-03 | 1.09E-02 |
| Maximum 2D Diameter Column | 2.28E+00 | 8.99E-01 | 1.14E-02 | 2.03E-02 |
| Maximum 2D Diameter Row | 1.26E+00 | 4.70E-01 | 7.45E-03 | 1.94E-02 |
| Maximum 2D Diameter Slice | -4.96E+00 | 1.06E+00 | 3.47E-06 | 4.51E-05 |

**Fig 7** We complement the curves from Figure 6 with the numerical results for the type 2 diabetes linear coefficient from the $\mu$ of the learned distributions from the GAMLSS model. Statistical significance after FDR correction is denoted in green.

We observed that MRI tended to yield smaller pancreas measurements than contrast and non-contrast CT. Additionally, we were able to reproduce the CT lifespan volumetry trends from manual measurements by Saisho et al., which together suggests that our AI pancreas segmentation approach may be potentially undermeasuring the pancreas on MRI[3]. This agrees with our manual quality assurance, in which pancreas segmentation failures occurred more often on MRI than CT. Additionally, our AI segmentation approach used TotalSegmentator, which is a previously trained model that is used for inference, meaning that it is predicting where the pancreas is located based on the distribution of the data that it was trained on[52,53]. Since CT is a quantitative modality, for which the intensities in each voxel are calibrated in Hounsfield units, our clinical data should be close in distribution to the training data. However, clinically acquired

MRI is typically not a quantitative modality, meaning that the intensity values in image voxels do not map back to physical properties in the way that is true for Hounsfield units (eg. there is no calibrated range of MRI intensities for fat across scans). Additionally in MRI, variations in acquisition parameters, scanner biases, and site effects hinder the generalization of AI models to unseen data. These challenges are compounded by the variability of MR contrast and the dependence of pancreatic tissue appearance on specific MR sequences, which complicate direct comparison with CT findings. Learning from multiple imaging modalities is the remarkably difficult harmonization challenge, with an entire field of research dedicated to the issue[59]. While our study suggests that AI models for pancreas segmentation may be reliably applied to CT imaging, their direct extension to MRI studies in future work should be approached with caution, unless controlled with robust harmonization strategies.

There is conflicting evidence in the literature on the volume of the pancreas in type 2 diabetes. Some studies found no difference in type 2 diabetes, and others state that the pancreas is significantly smaller in individuals with type 2 diabetes. Saisho et al. matched on sex, age, and BMI and found that the pancreas of people with type 2 diabetes was smaller than control[3]. Our findings agree with the evidence pointing towards a smaller pancreas in type 2 diabetes. Moreover, we demonstrate the shape of the pancreas is also altered in type 2 diabetes. While an additional investigation of fat infiltration across the lifespan could prove insightful, the vast majority of our CT had contrast, which renders fat estimation from Hounsfield units inaccurate.

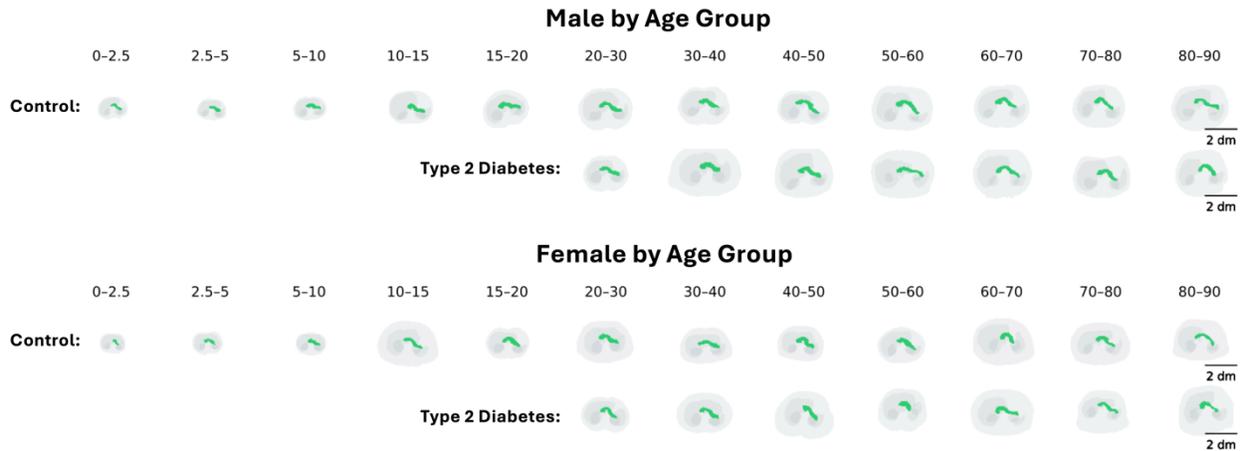

**Fig 8** Maximum intensity projections of pancreas segmentations with near-median sampled volumes by age group provide a visual reference for how the pancreas ages across the lifespan. These visualizations are to scale, with a scalebar in decimeters. While our earlier statistical tests detected a difference between type 2 diabetes and control, there was a large overlap in distributions. We reinforce that overlap in distributions here, since the pancreas looks similar in type 2 diabetes when using our projection of the relatively low resolution (3mm isotropic) image segmentations.

To compare subjects with different body sizes, studies often divide pancreas volume by body surface area (BSA), weight, or BMI—an approach sometimes used to assess differences between individuals with and without diabetes[23,35,36,37,38,39,40,41,42,43,44]. This approach is supported by demonstrations that there are correlations between pancreas volume and BSA, weight, and BMI[45,46,47,16]. In our comparison between patients with type 2 diabetes and non-diabetic controls, we accounted for body size variability by including weight as a covariate.

This work is limited in that we consider contrast in CT as a binary flag. In reality, contrast cycles through phases which impact the way the pancreas appears on imaging. Variability in contrast phase may influence pancreas segmentation via TotalSegmentator and may influence resulting measurements of the pancreas.

To advance our understanding of pancreas aging, future work should focus on quantifying changes in pancreatic size and shape using higher-resolution CT to enable finer structural analysis. In parallel, evaluating how anatomical aging patterns vary across clinical sites and populations will help establish robust baselines.

## 4 Conclusions

We find that in patients with type 2 diabetes, the pancreas is significantly smaller and different in shape than non-diabetic controls via a linear parameter for type 2 diabetes in the GAMLSS model. We provide reference trends for how control and type 2 diabetes pancreas morphology changes across the lifespan, as imaged in the hospital. We recommend that for AI analysis of large-scale clinical data, pancreas measurements should focus on CT, unless an MRI model is fine-tuned on the target data distribution. For cross-modality studies, it is essential to use robust image harmonization to ensure reliable pancreas measurements across imaging types. Without image harmonization, findings in one imaging modality may be difficult to interpret in the context of another.

*Acknowledgments*

This work was supported by Integrated Training in Engineering and Diabetes, grant number T32 DK101003 and MSTP T32: NIH NIGMS T32GM007347. This work used REDCap and VCTRS resources, which are supported by grant UL1 TR000445 from National Center for Advancing Translational Sciences, National Institutes of Health (NIH). This work received support from the Vanderbilt Diabetes Research and Training Center (DK020593), Division of Diabetes, Endocrinology, and Metabolic Diseases, and the Vanderbilt University Institute of Imaging Science Center for Human Imaging (1 S10OD021771 01). This work was supported by the Alzheimer's Disease Sequencing Project Phenotype Harmonization Consortium (ADSP-PHC) that is funded by NIA (U24 AG074855, U01 AG068057 and R01 AG059716). This work was supported by NSF career 1452485 and NSF 2040462. This research was funded by the National Cancer Institute (NCI) grant R01 CA253923-04, R01 CA 253923-04S1. This work was conducted in part using the resources of the Advanced Computing Center for Research and Education at Vanderbilt University, Nashville, TN. The Vanderbilt Institute for Clinical and Translational Research (VICTR) is funded by the National Center for Advancing Translational Sciences (NCATS) Clinical Translational Science Award (CTSA) Program, Award Number 5UL1TR002243-03. The content is solely the responsibility of the authors and does not necessarily represent the official views of the NIH. This work was supported by DoD grant HT94252410563. We extend gratitude to NVIDIA for their support by means of the NVIDIA hardware grant. Financial support was graciously provided by the National Institutes of Health (DK129979 and HD115565) and Breakthrough T1D (formerly JDRF) (1-INO-2023-1340-A-N). We gratefully acknowledge philanthropic support from Thomas J. and Karen K. Gentry. The study sponsors were not involved in the design of the study and did not impose any restrictions regarding the publication of the report. We have used AI as a tool in the creation of this content, however, the foundational ideas, underlying concepts, and original gist stem directly from the personal insights, creativity, and intellectual effort of the author(s). The use of generative AI serves to enhance and support the author's original contributions by assisting in the ideation, drafting, and refinement processes. All AI-assisted content has been carefully reviewed, edited, and approved by the author(s) to ensure it aligns with the intended message, values, and creativity of the work.

[26] Bagur, A. T., Ridgway, G., McGonigle, J., Brady, S. M. and Bulte, D., "Pancreas Segmentation-Derived Biomarkers: Volume and Shape Metrics in the UK Biobank Imaging Study," 131–142 (2020).

[27] Virostko, J., Craddock, R. C., Williams, J. M., Triolo, T. M., Hilmes, M. A., Kang, H., Du, L., Wright, J. J., Kinney, M., Maki, J. H., Medved, M., Waibel, M., Kay, T. W. H., Thomas, H. E., Greeley, S. A. W., Steck, A. K., Moore, D. J. and Powers, A. C., "Development of a standardized MRI protocol for pancreas assessment in humans," PLoS One **16**(8), e0256029 (2021).

[28] Syed, A.-B., Mahal, R. S., Schumm, L. P. and Dachman, A. H., "Pancreas Size and Volume on Computed Tomography in Normal Adults," Pancreas **41**(4), 589–595 (2012).

[29] Caglar, V., Songur, A., Yagmurca, M., Acar, M., Toktas, M. and Gonul, Y., "Age-related volumetric changes in pancreas: a stereological study on computed tomography," Surgical and Radiologic Anatomy **34**(10), 935–941 (2012).

[30] Kipp, J. P., Olesen, S. S., Mark, E. B., Frederiksen, L. C., Drewes, A. M. and Frøkjær, J. B., "Normal pancreatic volume in adults is influenced by visceral fat, vertebral body width and age," Abdominal Radiology **44**(3), 958–966 (2019).

[31] Zhou, Y., Lee, H. H., Tang, Y., Yu, X., Yang, Q., Kim, M. E., Remedios, L. W., Bao, S., Spraggins, J. M., Huo, Y. and Landman, B. A., "Multi-contrast computed tomography atlas of healthy pancreas with dense displacement sampling registration," Journal of Medical Imaging **12**(02) (2025).

[32] Sato, T., Ito, K., Tamada, T., Sone, T., Noda, Y., Higaki, A., Kanki, A., Tanimoto, D. and Higashi, H., "Age-related changes in normal adult pancreas: MR imaging evaluation," Eur J Radiol **81**(9), 2093–2098 (2012).

[33] Le Goallec, A., Diai, S., Collin, S., Prost, J.-B., Vincent, T. and Patel, C. J., "Using deep learning to predict abdominal age from liver and pancreas magnetic resonance images," Nat Commun **13**(1), 1979 (2022).

[34] Remedios, L. W., Liu, H., Remedios, S. W., Zuo, L., Saunders, A. M., Bao, S., Huo, Y., Powers, A. C., Virostko, J. and Landman, B. A., "Influence of early through late fusion on pancreas segmentation from imperfectly registered multimodal magnetic resonance imaging," Journal of Medical Imaging **12**(02) (2025).

[35] Al-Mrabeh, A., Hollingsworth, K. G., Steven, S. and Taylor, R., "Morphology of the pancreas in type 2 diabetes: effect of weight loss with or without normalisation of insulin secretory capacity," Diabetologia **59**(8), 1753–1759 (2016).

[36] Cai, Z., Chen, S., Chen, M., Meng, S., Hu, Y., Huang, M., Song, J., Huang, X., Yan, Z. and Liu, K., "Effects of diabetic ketoacidosis, islet cell function, and age of onset on pancreas size, morphology, and exocrine function in children with type 1 diabetes: an abdominal MRI study," J Endocrinol Invest (2025).

[37] WRIGHT, J. J., WILLIAMS, J. M., GREELEY, S. A. W., POWERS, A. C., VIROSTKO, J. and MOORE, D. J., "206-OR: Insulin Gene Mutation in Family with Monogenic Diabetes Leads to Smaller Pancreas," Diabetes **71**(Supplement_1) (2022).

[38] Pollé, O. G., Delfosse, A., Michoux, N., Peeters, F., Duchêne, G., Louis, J., Van Nieuwenhuyse, B., Clapuyt, P. and Lysy, P. A., "Pancreas Imaging of Children with Type 1 Diabetes Reveals New Patterns and Correlations with Pancreatic Functions," Pediatr Diabetes **2023**, 1–13 (2023).

[39] Goda, K., Sasaki, E., Nagata, K., Fukai, M., Ohsawa, N. and Hahafusa, T., "Pancreatic volume in type 1 und type 2 diabetes mellitus," Acta Diabetol **38**(3), 145–149 (2001).
23